\documentclass{article} % For LaTeX2e
\usepackage{iclr2021_conference,times}

\usepackage{amsmath,amsfonts,bm}

\def\eqref#1{equation~\ref{#1}}
\def\1{\bm{1}}

\DeclareMathAlphabet{\mathsfit}{\encodingdefault}{\sfdefault}{m}{sl}
\SetMathAlphabet{\mathsfit}{bold}{\encodingdefault}{\sfdefault}{bx}{n}

\usepackage{graphicx}

\title{What is Going on Inside Recurrent Meta Reinforcement Learning Agents?}

\author{Safa Alver \\
Mila - McGill University \\
\texttt{safa.alver@mail.mcgill.ca} \\
\And
Doina Precup \\
Mila - McGill University, DeepMind \\
}

\iclrfinalcopy % Uncomment for camera-ready version, but NOT for submission.
\begin{document}

\maketitle

\begin{abstract}
Recurrent meta reinforcement learning (meta-RL) agents are agents that employ a recurrent neural network (RNN) for the purpose of ``learning a learning algorithm''. After being trained on a pre-specified task distribution, the learned weights of the agent's RNN are said to implement an efficient learning algorithm through their activity dynamics, which allows the agent to quickly solve new tasks sampled from the same distribution. However, due to the black-box nature of these agents, the way in which they work is not yet fully understood. In this study, we shed light on the internal working mechanisms of these agents by reformulating the meta-RL problem using the Partially Observable Markov Decision Process (POMDP) framework. We hypothesize that the learned activity dynamics is acting as belief states for such agents. Several illustrative experiments suggest  that this hypothesis is true, and that recurrent meta-RL agents can be viewed as agents that learn to act optimally in partially observable environments consisting of multiple related tasks. This view helps in understanding their failure cases and some interesting model-based results reported in the literature.
\end{abstract}

\section{Introduction}
\label{sec:intro}

In recent years, meta reinforcement learning (meta-RL) agents have been reported to achieve significant results in terms of adapting to new but related tasks in a sample-efficient manner. After being trained on multiple tasks sampled from a pre-specified task distribution, in order to obtain the domain relevant priors, these agents were shown to be able to quickly adapt to unseen tasks sampled from the same distribution. In agents that employ recurrent neural networks (RNN)~\citep{duan2016rl, wang2016learning}, these priors correspond to the learned weights of the RNN, which are said to come to implement a learning algorithm that is very fast  and different from the learning algorithms  used to obtain these weights, e.g.\ policy gradient algorithms. However, due to their black-box nature, the underlying working mechanisms behind these agents are not yet understood and thus it is not clear what is going on inside the agent.

In this study, we shed light on the internal working mechanisms of these agents by first reformulating the meta-RL problem using the POMDP framework and then by hypothesizing that the RNN's learned activity dynamics is acting as a belief states for an agent acting in a partially observable environment (POE). To test this hypothesis, we perform several illustrative experiments on both bandit and gridworld tasks. Both our quantitative and qualitative results show that this hypothesis is true. This indicates that recurrent meta-RL agents can be viewed as agents that learn to act optimally in POEs consisting of multiple related tasks. The main contributions of this study are (1) the careful experiments that we perform and (2) the perspective that we provide on the internal working mechanisms of these agents. We hope that our findings will help in developing new meta-RL algorithms that can train agents to learn more powerful and interesting learning algorithms.

\textbf{Related Work.} Meta-RL is a long-standing problem that has been studied for more than three decades \citep[see e.g.][]{schmidhuber1996simple, thrun1998learning}. In recent years, with the popularization of deep learning,  many meta-RL studies have used deep neural networks to tackle this problem. The approaches used include learning a good weight initializiation for quick adaptation \citep{finn2017model}, an objective function for sample efficient training \citep{houthooft2018evolved} and hyperparameters for better asymptotic performance \citep{xu2018meta}. 

In this study, we are particularly interested in meta-RL algorithms in which the agents employ an RNN for the purpose of ``learning a learning algorithm'' that is capable of faster learning than regular, engineered learning algorithms \citep{duan2016rl, wang2016learning}. Unlike these studies however, we focus on understanding the underlying working mechanisms behind these agents, rather than on their performance against state-of-the-art engineered algorithms. The closest to our work is the study of \cite{ortega2019meta}, which aims for a theoretical understanding of recurrent meta-RL agents by recasting them within a Bayesian framework. However, unlike our study, they do not make explicit connections between recurrent meta-RL and the POMDP framework. Another closely related work is the study of \cite{humplik2019meta}, however, unlike our study, they do not provide explanations for the inner working mechanisms of recurrent meta-RL agents.

\begin{figure}
\includegraphics[width=1.00\linewidth]{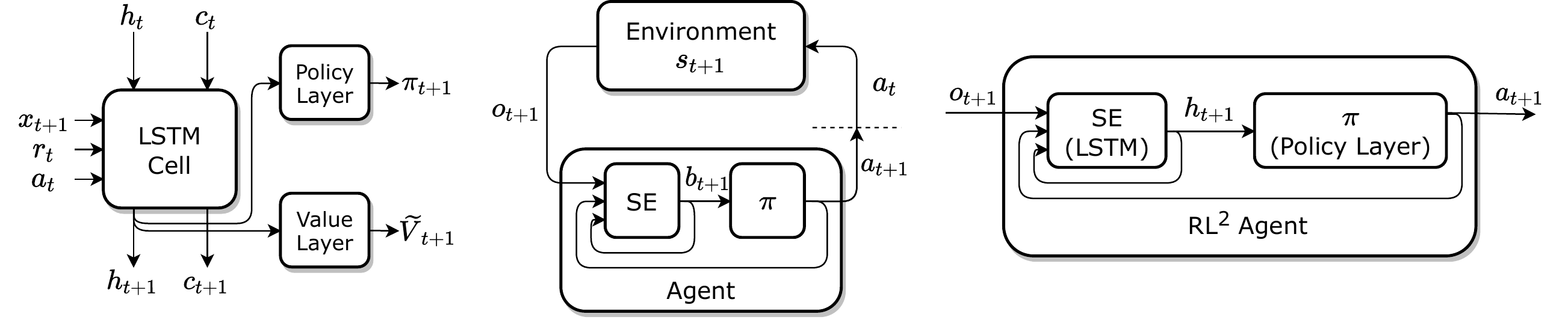}
\caption{(Left) The network architecture of the RL$^2$ agent. Here, the value layer is discarded after training. (Middle) The interaction between a regular agent and its POE, where the agent is composed of two parts: (1) a state estimator (SE) and (2) a policy ($\pi$). (Right) The inside view of the RL$^2$ agent: the LSTM acts as a state estimator and the policy layer acts a policy over belief states.}
\label{fig:pomdp}
\end{figure}

\section{Background}
\label{sec:back}

\textbf{Markov Decision Processes.} In RL \citep{sutton2018reinforcement}, the environment is usually modelled as a finite Markov decision process (MDP). A finite MDP is described as a tuple $\langle \mathcal{S}, \mathcal{A}, R, P, d_0, \gamma \rangle$, where $\mathcal{S}$ is the set of states (including the absorbing state), $\mathcal{A}$ is the set of actions, $R: \mathcal{S}\times\mathcal{A}\times\mathcal{S}\rightarrow \mathbb{R}$ is the reward function, $P:\mathcal{S}\times \mathcal{A}\times \mathcal{S}\rightarrow [0, 1]$ is the transition distribution, $d_0:\mathcal{S}\rightarrow [0,1]$ is the initial state distribution and $\gamma\in [0,1)$ is the discount factor. In the MDP framework, at each timestep $t$, after taking an action $a_t$, the agent receives the environment's next state $s_{t+1}$. The goal of the agent is to find a policy $\pi: \mathcal{S}\times \mathcal{A}\rightarrow [0,1]$ that maximizes the expected sum of discounted rewards $E_{\pi}[\sum_{t=0}^{\infty} \gamma^{t} R(S_t, A_t, S_{t+1}) | S_0 \sim d_0]$.

\textbf{Partially Observable MDPs.} A partially observable environment (POE) is modelled as a finite partially observable MDP \citep[POMDP, ][]{kaelbling1998planning}. A finite POMDP is described as a tuple $\langle \mathcal{S}, \mathcal{A}, R, P, d_0, \gamma, \Omega, O \rangle$, where $\mathcal{S}, \mathcal{A}, R, P, d_0$ and $\gamma$ are defined in the same way as in MDPs, $\Omega$ is the set of observations and $O:\mathcal{A}\times \mathcal{S}\times \Omega \rightarrow [0,1]$ is the observation distribution. At each timestep $t$, the environment is in some state $s_t$. The agent takes an action $a_t$, which causes the environment to transition to state $s_{t+1}$ with probability $P(s_t, a_t, s_{t+1})$. At the same time, the agent receives an observation $o_{t+1}$ with probability $O(a_{t}, s_{t+1}, o_{t+1})$ and an immediate reward $r_{t} = R(s_t, a_t, s_{t+1})$. Importantly, unlike in the MDP framework, the observation $o_{t+1}$ only gives partial information about the environment's current state $s_{t+1}$. Thus, in order to behave effectively, the agent has to keep an internal belief state $b_t \in \mathcal{B}$, a probability distribution over states that summarizes its previous experience up to timestep $t$. For this purpose, the agent employs a state estimator that computes a belief state $b_{t+1}$ based on the current observation $o_{t+1}$, previous action $a_{t}$, and previous belief state $b_{t}$. The observation $o_{t+1}$ may or may not contain $r_t$ \citep{izadi2005using}. In the case where $o_{t+1}$ contains $r_t$, the agent updates its belief state using  Bayes' rule as follows:
\begin{equation}
    b_{t+1}(s_{t+1}) = \eta \sum_{s_t\in \mathcal{S}} \left[ P(s_t,a_t,s_{t+1}) O(a_t,s_{t+1},o_{t+1}) P(r_t|s_t,a_t,s_{t+1}) \right] b_t(s_t),
    \label{eqn:belief_upd}
\end{equation}
where $\eta$ is the normalizing constant and $P(r_t|s_t,a_t,s_{t+1})=1$ if $R(s_t,a_t,s_{t+1})=r_t$ and $0$ otherwise. It should be noted that this update requires the agent's state estimator to have an accurate $P$, $O$ and $R$ model of the POMDP. The agent then uses this belief state $b_{t+1}$ for selecting the next action $a_{t+1}$ (see middle of Figure \ref{fig:pomdp}). The goal of the agent is to find a policy $\pi: \mathcal{B}\times \mathcal{A}\rightarrow [0,1]$ that again maximizes $E_{\pi}[\sum_{t=0}^{\infty} \gamma^{t} R(S_t, A_t, S_{t+1}) | S_0 \sim d_0]$.

\section{Problem Formulation}\label{sec:problem}

At a high level, in recurrent meta-RL \citep{duan2016rl, wang2016learning}, an agent is trained on an environment that consists of a finite set of tasks  sampled from a pre-specified task distribution. It is expected that the agent will learn a learning algorithm (a policy that first explores and then exploits) that can solve new tasks sampled from the same distribution. Each task is defined as a finite, episodic MDP and multiple episodic interactions are allowed with a given MDP. A single episodic interaction is called an episode and a sequence of episodic interactions is called a trial \citep{duan2016rl}. It is assumed that all the tasks share the same $\mathcal{A}$ and $\gamma$, but may differ in their $\mathcal{S}$, $R$, $P$ and $d_0$. However, this MDP formulation of the tasks can be problematic, because in its interaction with the environment, the agent does not receive any direct information on the identity of the current task, which is part of the environment's current state. Because of this, we reformulate the problem  as follows.

Let $\mathcal{T}$ be a set of tasks and $P(T)$ be a pre-specified distribution over the tasks $T\in \mathcal{T}$. We define each task $T$ as a finite episodic POMDP with a special structure, in which attempts to transition to the absorbing state instead result in a reset to its initial states for $K-1$ times in a row. Only in the $K$th attempt it results in a transition to the absorbing state, where $K>1$. We also define an interaction up to a reset as an episode and an interaction up to the absorbing state as a trial. As in the original formulation, we assume that the tasks in $\mathcal{T}$ share the same $\mathcal{A}$ and $\gamma$ but may differ in their $\mathcal{S}$, $R$, $P$, $d_0$, $\Omega$ and $O$. The objective under this formulation is to train the agent on a POE consisting of a finite set of tasks $\mathcal{T}^{\text{train}} = \{T_i \}, i = \{1,\dots,M\}$, where $T_i\sim P(T)$, so that the expected sum of discounted rewards on all the tasks in $\mathcal{T}^{\text{train}}$ is maximized. Here, in its training interaction with the POE, tasks are uniformly presented to the agent and at each timestep $t$, the agent only receives an observation $o_t$, rather than the POE's current state $s_t=(o_t,i_t)$ which contains the current task identity $i_t$. After being trained, the agent is expected to perform well on a new POE consisting of a new finite set of tasks $\mathcal{T}^{\text{test}} = \{T_j \}, j = \{1,\dots,N\}$, where $T_j\sim P(T)$, based on a small number of interactions with the tasks in $\mathcal{T}^{\text{test}}$. In this formulation, $\mathcal{T}^{\text{train}}$ and $\mathcal{T}^{\text{test}}$ may or may not share the same tasks, i.e., either $\mathcal{T}^{\text{train}} \cap \mathcal{T}^{\text{test}} \neq \emptyset$ or $\mathcal{T}^{\text{train}} \cap \mathcal{T}^{\text{test}} = \emptyset$.

\section{Experimental Results}
\label{sec:exp}

\textbf{Architectures and Algorithms.} We start this section by using RL$^2$ to collectively refer to the algorithms proposed by \cite{duan2016rl} and \cite{wang2016learning}. In RL$^2$, the meta-RL agent employs an LSTM \citep{hochreiter1997long} with a policy and value layer on top, and it is trained on a POE consisting of multiple related tasks, to map concatenated inputs of the form $x_{t+1}' = (x_{t+1}, r_{t}, a_{t})$ to policies $\pi(.|x_{t+1}')$ and value functions $V(x_{t+1}')$ (see left of Figure \ref{fig:pomdp}). Importantly, the agent's internal state, the LSTM's hidden activations and cell state $(h_t,c_t)$, is preserved between episodes in a trial, so the gathered information can be passed between episodes. 
After being trained to maximize the expected sum of discounted rewards over trials in the POE, the LSTM's activity dynamics is said to implement an efficient learning algorithm. However, due to the black-box nature of neural networks, it is not clear what is going on inside the agent.

Based on our POMDP formulation above, we hypothesize that the RL$^2$ agent's learned activity dynamics is acting as belief states for optimal behavior in its POE consisting of multiple related tasks. In this view, the agent's LSTM acts as a state estimator that receives current observations $o_{t+1} = (x_{t+1}, r_{t})$, previous actions $a_t$ and previous belief states $b_t = h_t$ and computes new belief states $b_{t+1} = h_{t+1}$, which are then used by the agent's policy layer to select next actions $a_{t+1}$ (see right of Figure \ref{fig:pomdp}). Here, $r_t$ in $o_{t+1}$ serve as partial information about the current task identity. In order to test this hypothesis, we prepare a training setting in which we first reset the RL$^2$ agent's internal state between episodes and then reveal the current task identity at each timestep as a part of the observation, after the first episode. This setting allows us to have a control over what information the agent is carrying in its activity dynamics between episodes in a trial. If an agent trained in this way displays a similar quantitative and qualitative performance to the RL$^2$ agent, we can safely conclude that the RL$^2$ agent's activity dynamics is acting as belief states on the current task identity. Since this agent does not have any means to implement a learning algorithm, we refer to it as RL$^1$.

We now describe our experiments for testing this hypothesis. For illustration purposes, we consider two simple environments consisting of only two tasks, on which we both train and test the agents. It should be noted that this kind of simplicity is required for performing informative experiments. More information on our network architectures and algorithm hyperparameters can be found in Appendix \ref{sec:A2_hyp}.

\textbf{(1) Dependent Bandit Environment.} We start by considering the dependent bandit environment proposed in \cite{wang2016learning}, which consists of structured two-arm bandit tasks in which the arm distributions are dependent Bernoulli distributions with parameters $(p, 1-p), p\sim [0,1]$. For illustration purposes however, here we consider a simpler version of this environment where $p\sim \{0,1\}$. That is, there are two tasks in the task distribution and in each task only one of the arms pays out a reward of $+1$, deterministically. Each arm pull is defined as an episode and the trials consist of $10$ episodes. In this environment, the agent receives only the current reward (which is either $0$ or $+1$) as an observation and the task identity is determined by the currently optimal arm.

\begin{figure}[t]
\includegraphics[width=0.28\linewidth]{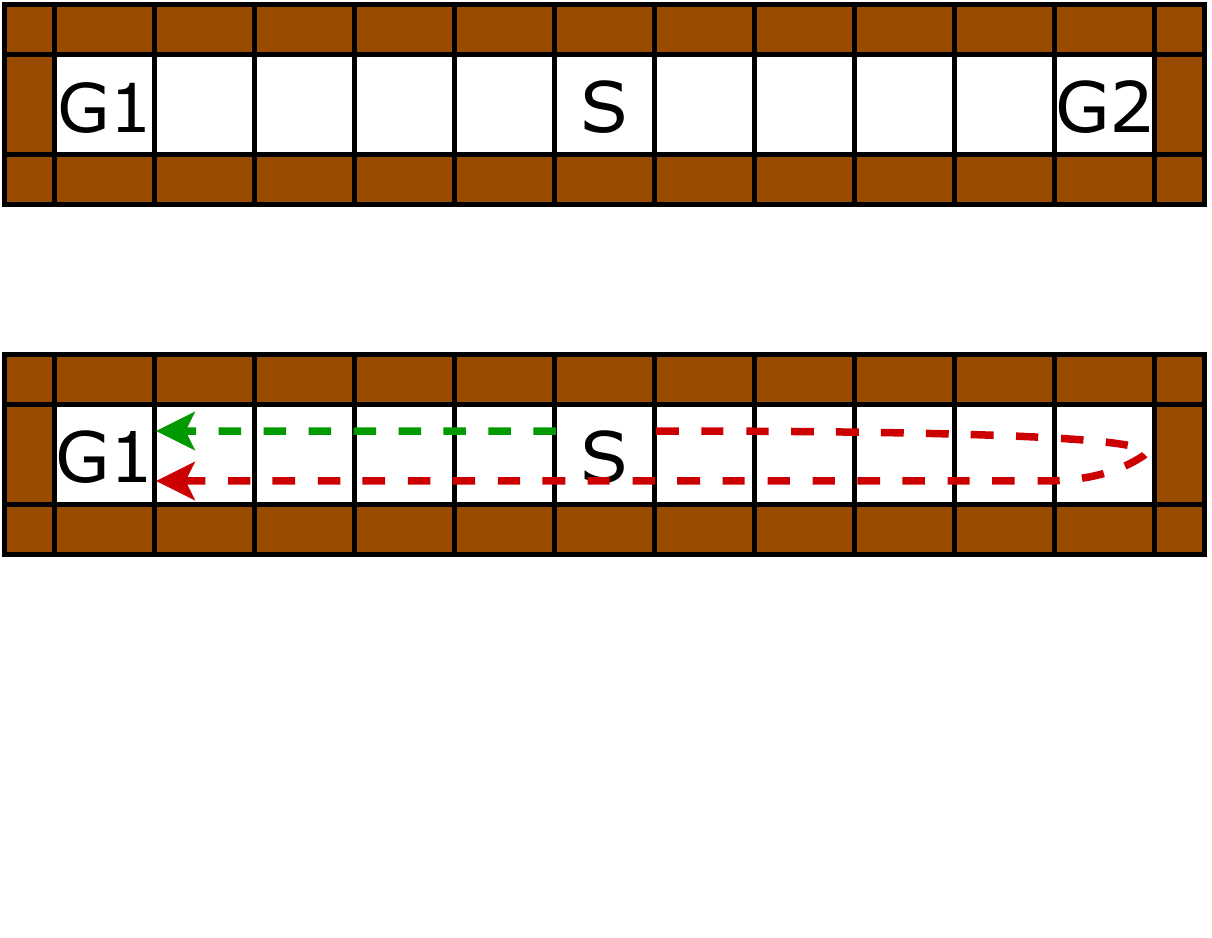} \hspace{0.05cm}
\includegraphics[width=0.35\linewidth]{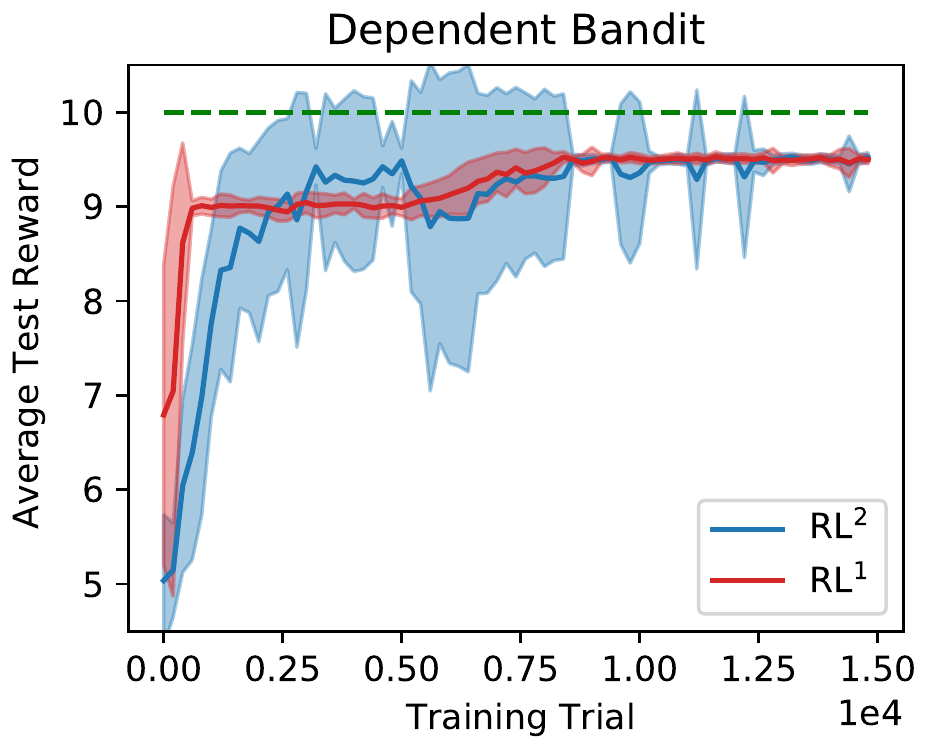}
\includegraphics[width=0.35\linewidth]{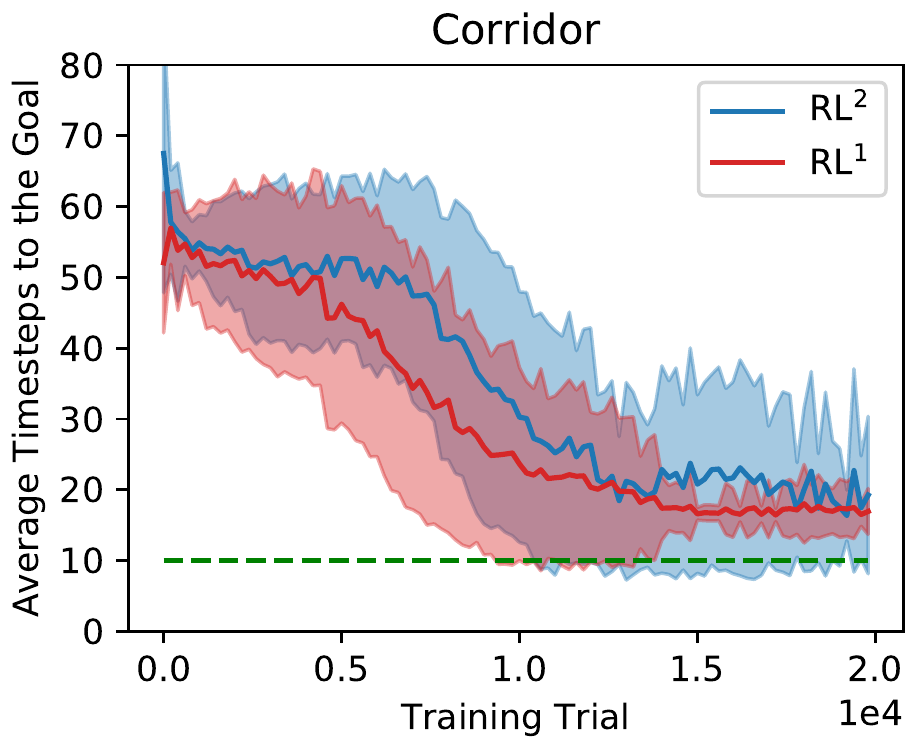}
\caption{(Left) The corridor environment and the behaviour of the agents when the goal is G1. In the first episode they both search for the goal following the red path. In the next one they directly navigate to it following the green one. (Middle) The performance of the agents on the dependent bandit environment. (Right) The performance of the agents on the corridor environment. The results are averaged over $100$ test time rollouts and $20$ independent training runs. The maximum performance, if the agent had known the optimal arm / goal location beforehand, is indicated with green dashed lines.}
\label{fig:results}
\end{figure}

We train both the RL$^2$ and RL$^1$ agents on this environment. The results are shown in the middle of Figure \ref{fig:results}. We observe that RL$^1$ shows similarity to RL$^2$ in the learning curves, asymptotic performance and in the final qualitative behavior, in which both choose a random arm in the first episode, and then choose the optimal arm in the following episodes, resulting in a final average test reward of $9.5$. This suggests that the information passed between the episodes in a trial is a belief state on the identity of the current task, which confirms our hypothesis that the activity dynamics is acting as belief states for optimal behavior in POEs consisting of multiple related tasks.

\textbf{(2) Corridor Environment.} In order to see if the previous results also generalize to more complex scenarios, we perform further experiments on the corridor environment. This environment is a gridworld that consists of a single corridor as shown in the left of Figure \ref{fig:results}. There are two tasks in which the agent spawns in state S and has to navigate either to the left (G1) or right (G2) end of the corridor. The task dynamics are deterministic and, after reaching the goal, the agent receives a reward of $+10$, which ends the episode. The trials consist of $2$ episodes. In this environment, the agent receives its own location (in one-hot form) and the current reward (which is either $0$ or $+10$) as an observation and the task identity is determined by the current location of the goal.

We train both the RL$^2$ and RL$^1$ agents on this environment. The results are shown in the right of Figure \ref{fig:results}. We again observe that RL$^1$ shows similarity to RL$^2$ in the learning curves, asymptotic performance and final qualitative behavior. Both agents learn to search for the goal in the first episode, and then directly navigate to it in the second one (see the left of Figure \ref{fig:results}) resulting in a final average timesteps to the goal of $15.5$. This again suggests that the information passed between episodes is a belief state on the identity of the current task and further confirms our hypothesis.

\section{Conclusion and Discussions}

To summarize, in this study, we shed light on the internal working mechanisms of recurrent meta-RL agents by first reformulating the meta-RL problem using the POMDP framework and then by performing experiments to confirm the hypothesis that the learned activity dynamics of these agents is acting as belief states. It should be noted that even though we have specifically considered recurrent meta-RL agents that employ an RNN, our results naturally apply to meta-RL agents that employ any kind of memory-based architecture. This suggests that recurrent meta-RL agents can be viewed as agents that learn to act optimally in POEs consisting of multiple related tasks.

This view helps in understanding the failure cases recently reported in the literature. For instance:
\begin{itemize}
    \item The experiments of \cite{humplik2019meta} show that the final qualitative behavior of the RL$^2$ agent does not change when its input is augmented with the outputs of a belief network.  In our view, this is expected, as the RL$^2$ agent is already employing a state estimator inside.
    \item \cite{alver2020brief} have reported that the RL$^2$ agent performs poorly when the rewards are sparse. This is also expected, because when no immediate reward is present, the agent's state estimator cannot update the belief states to infer the current task identity.
    \item \cite{houthooft2018evolved} have reported that after being trained to navigate a simulated ant robot in the forward direction, the RL$^2$ agent cannot navigate it to the opposite direction. In the view that we provide, this is expected, as navigating backwards is not a task that the agent was trained on, thus its state estimator cannot correctly compute the belief states for effective behavior.
    \item The experiments of \cite{wang2021alchemy} show that if the current task identity is not provided during training, a memory-based meta-RL agent (similar to the RL$^2$ agent) performs similarly to a random policy on the Alchemy benchmark. This is expected, as the complexity of the Alchemy benchmark prevents the agent from forming an accurate state estimator that can correctly compute the belief states.
    \item \cite{yu2020meta} have reported that the RL$^2$ agent performs poorly even in the simplest evaluation mode of the Meta-World benchmark. Similar to the case with \cite{houthooft2018evolved}, this is expected, as the test tasks in this benchmark are tasks on which the agent was not trained.
\end{itemize}

Our view also helps in explaining the interesting model-based behavior of the RL$^2$ agent reported by \cite{wang2016learning}. Even though the RL$^2$ agent was trained by a model-free algorithm, it displays model-based behavior. In the view that we provide, this is expected, as in order for the agent's state estimator to compute accurate belief states, it also has to model the transition distribution $P$, the observation function $O$ and the reward function $R$ of the current task in the POE (see Eqn.\ \ref{eqn:belief_upd}).

In future work, we hope to use our improved understanding provided in this paper in order to develop better algorithms that can train agents to learn more powerful and interesting learning algorithms.

\bibliography{iclr2021_conference}
\bibliographystyle{iclr2021_conference}

%\newpage

\appendix
\section{Appendix}

\subsection{Network Architecture and Hyperparameters}
\label{sec:A2_hyp}

In our experiments we used the same network architecture for both the bandit and gridworld tasks. We first concatenate the current observation (containing both the current observation and the current reward) with the agent's previous action (in one-hot form) and then pass it to an LSTM cell with $48$ units. We then pass the LSTM cell's output through two fully connected layers to obtain the policy logits and the value function. All the weights are initialized to zero before the training process.

The hyperparameters of the RL$^2$ and RL$^1$ agents, which are built on top of the A2C algorithm, are given in Table \ref{tab:hyper}.

\begin{table}[h!]
    \caption{Hyperparameters of the RL$^2$ and RL$^1$ agents in the (left) bandit and (right) gridworld tasks.}
    \centering
    \begin{tabular}{l|r}
        \hline
        Learning rate & $1e-3$ \\
        Discount & $0.80$ \\
        Entropy coefficient & $0.001$ \\
        Gradient clip & $1.00$ \\
        \# of episodes in trial & 10 \\
        Value function loss coefficient & $0.05$ \\
        \hline
    \end{tabular}
    \hspace{0.5cm}
    \begin{tabular}{l|r}
        \hline
        Learning rate & $1e-4$ \\
        Discount & $0.90$ \\
        Entropy coefficient & $0.01$ \\
        Gradient clip & $5.00$ \\
        \# of episodes in trial & 2 \\
        Value function loss coefficient & $0.05$ \\
        \hline
    \end{tabular}
    \label{tab:hyper}
\end{table}

\end{document}